\documentclass[10pt, a4paper]{article}
\usepackage{lrec}
\usepackage{multibib}
\usepackage{graphicx}
\usepackage{tabularx}
\usepackage{soul}
\pdfoutput=1
\usepackage[implicit]{hyperref}
\usepackage{epstopdf}
\usepackage[latin1]{inputenc}
\usepackage{hyperref}
\usepackage{xstring}

\title{A System for Automated Image Editing from Natural Language Commands}

\name{Jacqueline Brixey, Ramesh Manuvinakurike, Nham Le, Tuan Lai, Walter Chang, Trung Bui}

\address{Adobe Research}

\abstract{
This work presents the task of modifying images in an image editing program using natural language written commands. We utilize a corpus of over 6000 image edit text requests to alter real world images collected via crowdsourcing. A novel framework composed of actions and entities to map a user's natural language request to executable commands in an image editing program is described. We resolve previously labeled annotator disagreement through a voting process and complete annotation of the corpus. We experimented with different machine learning models and found that the LSTM, the SVM, and the bidirectional LSTM-CRF joint models are the best to detect image editing actions and associated entities in a given utterance.  %\\ \newline \Keywords{keyword1, keyword2,keyword3} 
}

\begin{document}

\maketitleabstract

\section{Introduction}

The need to edit photographs has existed for as long as there has been photography. Cameras inherently have limitations in capturing all real world lighting and colors, causing users to correct these limitations through later editing. Digital image processing programs, such as Photoshop, made the process of image editing more accessible. However, novice users often find that they need significant training in order to successfully carry out desired edits, illustrated on web sites such as Reddit's Photoshop Request forum\footnote{https://www.reddit.com/r/PhotoshopRequest/} and Zhopped\footnote{http://zhopped.com/} where users submit image edit requests. They often communicate their editing needs using ordinary, non-technical language, such as:

\begin{itemize}
  \setlength{\parskip}{0ex}
  \setlength{\itemsep}{0ex}
  \item There is a spot on my wedding dress. Can someone please remove it. Please!
  \item He just passed away. He`d want his obituary photo to look phenomenal, but I think the lighting on his face is bad. Can someone fix that for me please?
\end{itemize}

%At Adobe Research,  // Anonymous submission
We aim to develop a software tool that will assist all users to achieve their image editing goals by interpreting and executing natural language image edit requests. This tool will allow users to independently manipulate images without the assistance of an expert user, and will not require learning technical vocabulary. Our work is a first step towards developing such a tool. We utilize our previous work, the Edit Me corpus \cite{manuvina-EtAl:2018:LREC}, a data set of written edit requests to alter real world images and related framework. We contribute to the data set by resolving previous annotation discrepancies and completing the unlabeled annotations. All utterances were annotated using our framework developed in Edit Me, a mapping of requests to executable commands in image editing software. We then implemented a two-level system, in which the first level classifies actions in an utterance, and the second level identifies relevant properties related to the action.

\section{Image and Edit Request Research}
\subsection{Previous Research}
Research combining vision and language include systems in visual question answering \cite{antol2015vqa}, visual storytelling \cite{visualstorytelling2016}, generating questions about an image \cite{mostafazadeh2016}, and question-answer interactions grounded on information shown in an image \cite{mostafazadeh2017}. Previous work to understand image descriptions \cite{kulkarni2013babytalk} is essential to our work, as illustrated in the forum examples above (\emph{lighting on his face}). Our work also draws on work to identify visual references (\emph{on my wedding dress}) \cite{maikeeveagent,de2016guesswhat}. In \cite{laput2013pixeltone}, a mobile interface for users to edit images through spoken language was developed, and is the only known previous work on image editing. They employed a rule-based system; we expand on this knowledge by handling a larger variety and structure of natural language image editing utterances through our machine learning implementations.

\subsection{Corpus}
We utilize our Edit Me corpus of 9101 text edit requests (44727 word tokens) that was created using Amazon Mechanical Turk\footnote{https://mturk.com} crowd-workers (called turkers for the rest of this work) \cite{manuvina-EtAl:2018:LREC}. The elicited requests illustrated a wide and challenging variation in vocabulary, utterance structure, and domain knowledge to accomplish similar editing outcomes. For vocabulary, similar but distinct terms were used to execute similar actions, such as crop, cut out, and delete to alter the dimensions of an image. While the majority of requests began with an imperative verb, for example \emph{Crop the left side of the image}, other prevalent utterances used modal verbs and formed requests as a comment, such as \emph{The image is blurry}. A lack of domain knowledge led to some ambiguous cases, such as the use of ``zoom in'' indicating to bring a portion of the photo into closeup, but could also indicate a request to crop (ex. \emph{Zoom in on the zebra}).

\subsection{Framework for Image Edits}
Our annotation framework serves as an intermediary language that interprets requests in terms of editing software functionality. 

An Image Edit Request (IER) contain an action that could be completed by an image editing program. IERs are composed of at most one action, and zero or more related entities. The framework maps between an explicit or implicit word or a phrase to one of 18 actions: \emph{adjust, delete, crop, add, replace, apply, zoom, rotate, transform, move, clone, select, swap, undo, merge, redo, other,} and \emph{scroll}. While actions provide a first level of understanding of an IER, entities complete the interpretation of how an action should be applied. The framework supports five types of entities: \emph{attribute, modifier/value, object, region,}  and \emph{intention}. 

\begin{figure}[h]
\centering
  \includegraphics[scale=0.58]{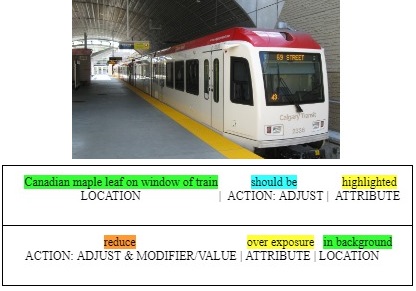}
  \caption{Example image from the corpus with annotated IERs.}
  \label{fig:annotateexample}
\end{figure}

The framework`s flexibility permits for multiple annotations of the same entity type in a single IER. It also supports an utterance having no entities, which occurred in 3\% of the instances in the data set. A unique feature of the framework is that the same word in an utterance can have multiple labels or one can be a subset of another. In the example \emph{Increase the saturation}, the word \emph{increase} is annotated as an \emph{adjust} action and as a \emph{modifier/action} entity. 

\section{Annotation}
The annotation scheme was previously tested for inter-rater reliability on a sample of 600 utterances (results shown in Table \ref{interrater}) \cite{manuvina-EtAl:2018:LREC}. The highest agreement was attained for the action types; agreement on entities was lower but still well above chance. However, for some entities, namely \emph{modifier/value}, agreement borders on chance level. 

\begin{table}[h]
\begin{center}
\begin{tabular}{|l|lll|}
\hline Feature & \multicolumn{3}{|c|}{Krippendorff`s alpha} \\\hline
IER vs. comment &0.28& 0.53& 0.35 \\
Action type& 0.74 &0.62 &0.59\\
Attribute& 0.47 &0.41 &0.38 \\
Object& 0.51& 0.27& 0.47\\
Region &0.55& 0.35& 0.43 \\
Modifier/value& 0.31 &0.04 &0.07 \\
Intention &0.51 &0.67 &0.52 \\
\hline
\end{tabular}
\end{center}
\caption{\label{interrater} Inter-rater reliability for 3 groups of 3 annotators. }
\end{table}

Due to the low levels of agreement, we determined that the annotations should be redone by an annotator with additional training and support before computational modeling could be attempted. Furthermore, not all of the corpus was annotated by highly trained annotators, rather by Turkers that showed even lower levels of inter-annotator agreement (see \cite{manuvina-EtAl:2018:LREC} for these scores). All utterances were thus annotated by our annotator who was trained by reviewing and discussing annotations and annotator disagreements in the previous labels. 

The most common action represented in the data set was \emph{adjust} (comprising 44\% of actions in the corpus). The framework was designed for interactive dialogue, not only the single instance IERs elicited in the crowd-sourced corpus. As a result, actions like \emph{undo, redo, select, merge,} and \emph{scroll} are infrequent or do not occur in the corpus. For entities, \emph{attribute} occurred in 56\% of the annotated utterances; \emph{modifier/value} was labeled in 32\% of IERs; \emph{object} was labeled in 30\% of utterances; \emph{region} was annotated in 60\% of the IERs; and \emph{intention} occurred in 29\% of the relabeled data. 
%\textbf{explain changes in numbers}

%\subsection{Resolving Action Annotation Discrepancies}
%To resolve action type discrepancies between previous annotators, who completed 600 annotations, and the new annotator who labeled the entire corpus, the action that the majority of the three trained annotators agreed upon was used. In the case that annotators differed on whether the utterance was a comment or an IER, we used the IER annotation. We propose that allowing the system to interpret implicit requests, such as the previous example of \emph{The photo should be brighter}, will produce a more robust system that can handle a larger variety of vocabulary and syntax. 

%In the case that all annotators in the group had differing annotations on the same utterance, we employed a voting system. Three trained annotators reviewed annotations that were different, approximately 200 of the total 6000 annotations, or 3\% of the total corpus, that were annotated. The action label that received the majority of the votes was the label used in the finalized data set for training and testing the machine learning models. 

\section{Methods}
\subsection{Preprocessing}
The corpus of annotated utterances was first filtered for executable actions. In this stage, we removed all utterances without an IER, such as in \emph{This image should have been taken with a Nikon}. Utterances with an \emph{other} action were also filtered out (0.01\% of the corpus). IERs labeled \emph{other} contained some level of ambiguity that made the requested edit impossible to execute, such as \emph{Clean up the pavement}. It is unclear if the user would be satisfied by the pavement being edited to all be a uniform color, or perhaps by deleting foliage on the pavement. We leave the investigation of this particular action for future work.  

To prepare the model input, Glove \cite{pennington2014glove} was selected to map the IERs to vectors. Annotated entity sequences were converted to BIO (beginning-in-out) sequences. For example, the utterance \emph{Crop the image}, annotated as \emph{ [IER :  [ACTION-CROP : crop ] [LOCATION : the image ] ]}, would become O, B-LOCATION, I-LOCATION. Nested entities, such as in \emph{Add a warmer hue}, where ``warmer'' is labeled with both the \emph{attribute} and \emph{value} labels, presented important considerations for the BIO encoder since nesting with a high degree of depth is possible. Nested entities account for 4\% of all the entities in the corpus, hence it was possible that a nesting depth that occurred in the testing data set did not occur in the training. Both models investigated in this work often fail when encountering a novel nested entity beyond the depth seen in training. However, using the innermost entity would allow the image editing system to still respond to a novel multi-level nested utterance, albeit with an incomplete outcome. As all labels carry the same amount of importance, and there was no annotation order rule for nested labels, we expected that performance would be similar for any depth of nesting ultimately used. For these reasons, we arbitrarily selected to use the innermost label. 

Finally, fixed sets of utterances for training and testing were created by randomly selecting utterances from the corpus. For actions, the training set contained 4958 utterances (75\% of the corpus) and 1584 utterances for testing. The entities data was split into training (80\% of the corpus), validation (10\%), and testing (10\%). 

\subsection{Structure of the Model}
Our predictive model is composed of two levels. The first level classifies only actions in an IER. The results of the classification process are then passed to a second level which detects sequences of only the entities. We propose that splitting the model will encourage filtering out IERs with ambiguous executable actions, thus preventing these utterances from being processed further.  

In the first level to classify actions, we evaluate a state-of-the-art Long Short Term Memory (LSTM) model with Tensor Flow\footnote{https://www.tensorflow.org/} as the backend against three baseline algorithms: Support Vector Machine (SVM), Logistic Regression, and Random Forest. All baseline machine learning models were implemented in Python using Scikit Learn\footnote{http://scikit-learn.org/stable/}. 
%In detail, SVM is a Linear SVR model with a loss function of ``squared hinge'', a default limit of 1000 iterations, and a default tolerance criteria for stopping value of 0.0001. For Logistic Regression, we implemented it with a l1 penalty and a tolerance criteria for stopping of 1e-6. Random Forest used a max depth of two and a zero random state. The LSTM model was implemented with one LSTM layer with 128 hidden units, a learning rate of 0.0001, and trained on 10,000 iterations. 

In the second level to detect entities in an IER, we compared Conditional Random Fields (CRF) with default parameters (namely, the L-BFGS training algorithm) in Scikit Learn as a baseline against a state-of-the-art model, BiLSTM-CRF \cite{lample2016neural}. BiLSTM-CRF combines a bidirectional LSTM with a CRF model. Previous experiments indicated the ability of this model to improve upon the limitations of CRFs by constraining the independence of output labels via the LSTM component. 
%The model first passes an utterance through a bidirectional LSTM model, which yields a matrix of scores for potential sequences for the utterance. Next a softmax is computed over all possible label sequences from the LSTM's matrix of scores to produce a probability for each possible sequence in the matrix. Training maximizes the probabilities of correct sequences, making valid tag sequence probabilities for a test utterance more likely to be the output of the CRF step of the model. 
We utilized the default parameters for BiLSTM-CRF.
%that uses a single layer for the forward and backward LSTMs with a dropout rate to 0.5. 

%\subsection{Synthetic Data}
%The collected data set is highly skewed, as evidenced in the frequencies of occurrence in Table \ref{axnframework}. Preliminary machine learning models performed well by simply classifying all utterances as the ADJUST action, but performed poorly on classifying rarer classes, such as CLONE. We also found that initial versions of the sequence-to-sequence LSTM model required substantially more data than available in the image edit requests corpus. Thus, we generated synthetic data from examples in the corpus to both balance action categories and efficiently create a larger training corpus.

%\textbf{@Ramesh please add details in this section} The synthetic data is generated by utilizing the skeleton of annotations in an utterance, eg. leaving  IER, action: adjust, and attribute in the utterance \emph{Increase the saturation}. The model builds lists of words that occurred in each annotation slot. Utilizing skeletons from real utterances, the model randomly chooses a word to place in each annotation slot in the skeleton to generate novel utterances (resulting example). 

\section{Results}

\subsection{Action}
\begin{figure}[h]
\centering
  \includegraphics[scale=0.22]{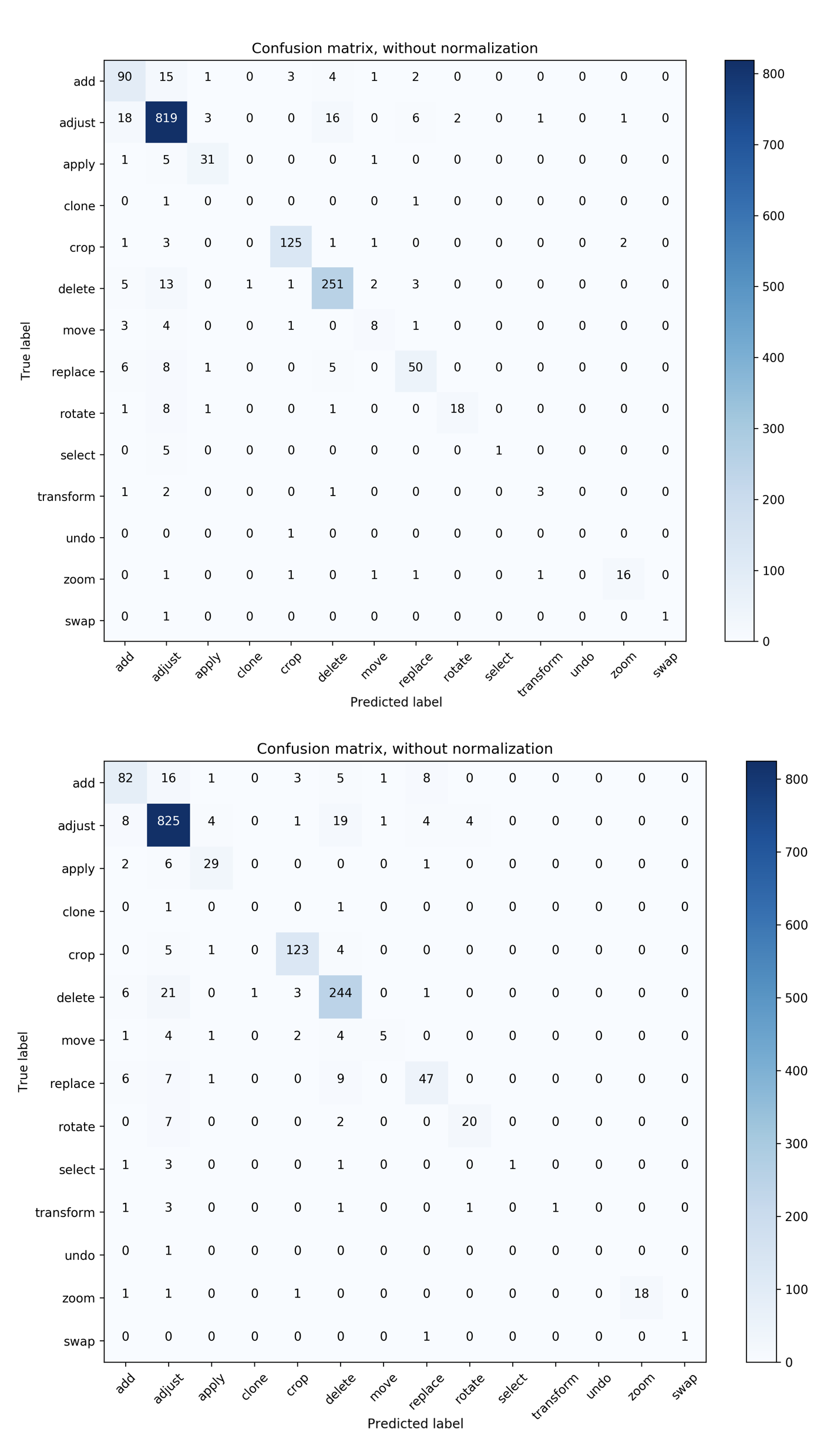}
  \caption{Confusion matrix for SVM (top) and LSTM (bottom)}
  \label{confusion}
\end{figure}

The best F1 score reported for this task was given by the LSTM and SVM models (Table \ref{action}). One concern was that the highly skewed data set would present problems, namely that all utterances would be classified as the majority class, the \emph{adjust} action. As the confusion matrix in Figure \ref{confusion} attests, however, the the LSTM model (shown on the top) correctly classified minority classes with high accuracy, for example, the \emph{rotate} action was correctly classified most of the time despite its low frequency of occurrence in the data. The SVM algorithm (confusion matrix shown on the bottom of Figure \ref{confusion}) was not as robust at correctly classifying the majority label, but did perform better than LSTM at predicting the three next largest classes (\emph{add, crop, delete}).

\begin{table}[h]
\begin{center}
\begin{tabular}{|l|c|}
\hline \bf   & F1 Score \\
\hline 
\bf Logistic Regression & 0.87 \\
\bf SVM & \textbf{0.89} \\
\bf Random Forest &  0.58 \\
\bf LSTM & \textbf{0.89} \\
%Synthetic 10k and testing on Synthetic 10k &0.89  &0.87 &0.28 & 0.86\\
%Synthetic 10k and testing on Real &0.40  &0.37 &0.12 & 0.37\\
%Synthetic 20k and testing on Synthetic 10k &0.90  &0.88 &0.26 & 0.89\\
%Synthetic 20k and testing on Real &0.40  &0.38 &0.08 & 0.38\\
%Synthetic 50k and testing on Synthetic 10k  &0.91  &0.90 &0.29 & 0.88 \\
%Synthetic 50k and testing on Real  &0.41  &0.37 &0.05 & 0.40\\
%Synthetic 80k and testing on Synthetic 10k &0.91  &0.91 &0.26 &  0.89 \\
%Synthetic 80k and testing on Real &0.41  &0.38 &0.01 & 0.43\\
%Synthetic 80k+Real and testing on Real &0.78  &0.78 &0.01 & 0.76\\
\hline
\end{tabular}
\end{center}
\caption{\label{action} Accuracy for each machine learning algorithm to classify the action label.}
\end{table}

\subsection{Entities}
For entities, we experimented with producing only the innermost entity as well as nested entities. Table \ref{entity} gives the results for correctly translating an utterance into a sequence of executable entities. 
\begin{table}[h]
\begin{center}
\begin{tabular}{|l|cc|}
\hline \bf  Entity Structure & \bf CRF &  \bf BiLSTM-CRF  \\ \hline
Nested entities &0.68 & 0.68 \\
Only innermost entities &0.66 & \textbf{0.73} \\
%Training on Synthetic 80k + Real and testing on Real&0.22 & 0.67\\
%Training on Synthetic 80k and testing on Real & 0.24 & 0.39\\
%Training and testing on Synthetic 100k& 0.27 & 0.94\\
\hline
\end{tabular}
\end{center}
\caption{\label{entity} F1 scores for entity labels. }
\end{table}

The results indicate that the state-of-the-art algorithm, BiLSTM-CRF, performs substantially better than the baseline CRF model for innermost entities. This indicates that the constraints induced by the BiLSTM component of the BiLSTM-CRF have a meaningful effect on the sequencing capability of the overall model. 

%Table \ref{entity} also shows that the performance decreases for the baseline CRF algorithm when only the innermost entities are used. This is unusual as the complexity of sequencing is reduced when using only the innermost entities, it can be concluded that the CRF model benefits from the depth of nested entities. It is evident then that the BiLSTM component of the BiLSTM-CRF model is responsible for the improvement in accuracy when using only the innermost entities. 

\section{Conclusions and future work}
This paper provided first steps towards automated image editing communicated through natural language. We contributed to the Edit Me corpus by annotating the remainder of the corpus and by re-annotating utterances with previous annotation disagreement.
%We explored the use of synthetic data to both correct for unbalanced data and to augment the number of training examples. 
We also evaluated a two-level system to classify actions and sequence entities in an edit request. We determined that the SVM model performed as well as LSTM for classifying actions, and that BiLSTM-CRF performs better at sequencing of only the innermost label of nested entities than the baseline learning algorithm. In future work, we plan to investigate a joint model that can predict both actions and entities. In addition, the two-level action and entities model will be applied to image editing dialogues to explore transfer learning. Finally, in many cases, entities require further parsing before being fully executable. We leave it to future work to parse vague entities.

% \nocite{*}
\section{Bibliographical References}
\label{main:ref}

\bibliographystyle{lrec}
\bibliography{xample}

%\section{Language Resource References}
%\label{lr:ref}
%\bibliographystylelanguageresource{lrec}
%\bibliographylanguageresource{xample}

\end{document}